\documentclass[runningheads]{llncs}
\usepackage{graphicx}

\usepackage{tikz}
\usepackage{comment}
\usepackage{amsmath,amssymb} 
\usepackage{color}

\usepackage[accsupp]{axessibility}  

\usepackage{makecell}
\usepackage{graphicx}
\usepackage{amsmath, bm}
\usepackage{tabularx,booktabs}
\newcolumntype{C}{>{\centering\arraybackslash}X} 
\usepackage{adjustbox}
\usepackage{multirow}
\usepackage{graphicx,xcolor,colortbl}
\definecolor{LightCyan}{rgb}{0.88,1,1}
\definecolor{Redx}{rgb}{0.9,0.61,0.60}

\usepackage{subfigure}
\usepackage{marvosym}

\begin{document}
\pagestyle{headings}
\mainmatter
\def\ECCVSubNumber{0000}  

\title{Unsupervised Domain Adaptation for Monocular 3D Object Detection via Self-Training} 

\titlerunning{STMono3D}
%
\author{Zhenyu Li\inst{1} \and
Zehui Chen\inst{2} \and
Ang Li\inst{3} \and
Liangji Fang\inst{3} \and
Qinhong Jiang\inst{3} \and \\
Xianming Liu\inst{1} \and
Junjun Jiang\inst{1}\textsuperscript{\Letter}}
\authorrunning{Z. Li et al.}
%
\institute{Harbin Institute of Technology \and
University of Science and Technology of China \and
SenseTime Research\\
\email{\{zhenyuli17, csxm, jiangjunjun\}@hit.edu.cn lovesnow@mail.ustc.edu.cn} \email{\{liang1, fangliangji, jiangqinhong\}@senseauto.com}}
\maketitle

\begin{abstract}

    Monocular 3D object detection (Mono3D) has achieved unprecedented success with the advent of deep learning techniques and emerging large-scale autonomous driving datasets. However, drastic performance degradation remains an unwell-studied challenge for practical cross-domain deployment as the lack of labels on the target domain. In this paper, we first comprehensively investigate the significant underlying factor of the domain gap in Mono3D, where the critical observation is a depth-shift issue caused by the geometric misalignment of domains. Then, we propose \textit{STMono3D}, a new self-teaching framework for unsupervised domain adaptation on Mono3D. To mitigate the depth-shift, we introduce the \textit{geometry-aligned multi-scale} training strategy to disentangle the camera parameters and guarantee the geometry consistency of domains. Based on this, we develop a teacher-student paradigm to generate adaptive pseudo labels on the target domain. Benefiting from the end-to-end framework that provides richer information of the pseudo labels, we propose the \textit{quality-aware supervision} strategy to take instance-level pseudo confidences into account and improve the effectiveness of the target-domain training process. Moreover, the \textit{positive focusing training} strategy and \textit{dynamic threshold} are proposed to handle tremendous FN and FP pseudo samples. STMono3D achieves remarkable performance on all evaluated datasets and even surpasses fully supervised results on the KITTI 3D object detection dataset. To the best of our knowledge, this is the first study to explore effective UDA methods for Mono3D.

\end{abstract}
 
\section{Introduction}
Monocular 3D object detection (Mono3D) aims to categorize and localize objects from single input RGB images. With the prevalent development of cameras for autonomous vehicles and mobile robots, this field has drawn increasing research attention. Recently, it has obtained remarkable advancements~\cite{chen2016mono3d,brazil2019m3drpn,xu2018multifusion,wang2021fcos3d,wang2022pgd,park2021dd3d,reading2021monoflex} driven by deep neural networks and large-scale human-annotated autonomous driving datasets~\cite{geiger2012kitti,caesar2020nusc,kesten2019lyft}.

\begin{figure}[t]
    \centering
    \footnotesize
    \scalebox{0.95}{%
    \begin{tabular}{l}
        \includegraphics[width=1\linewidth]{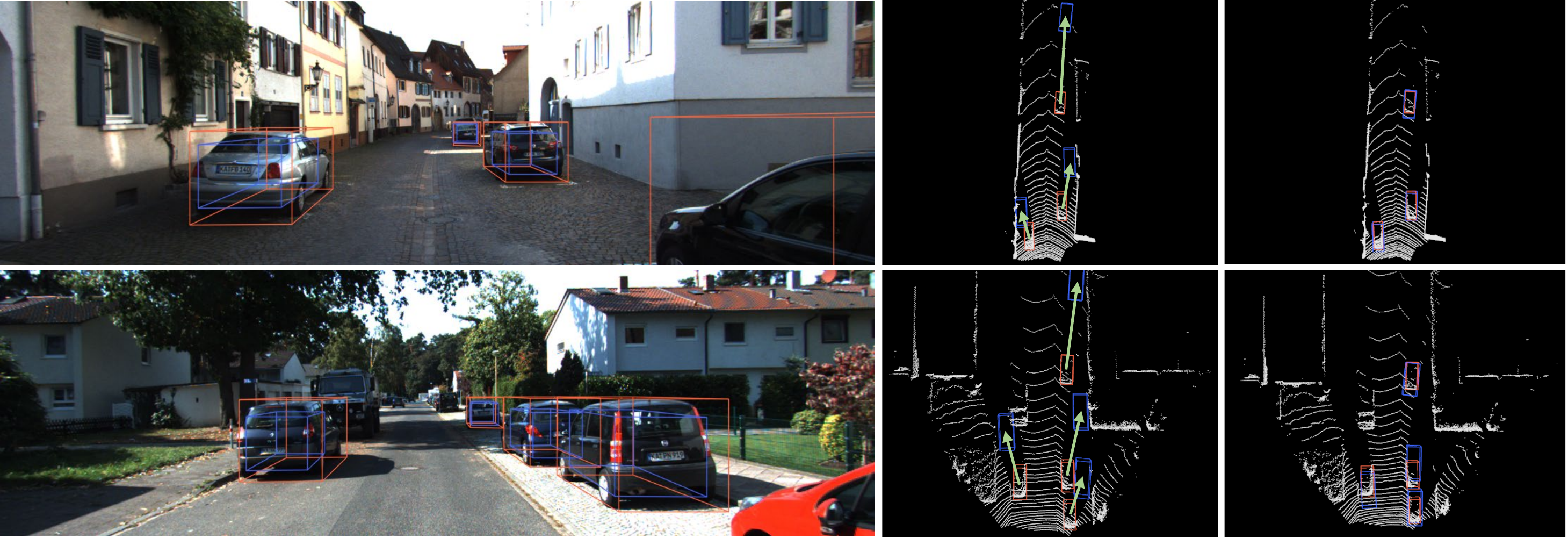}\\
        \hspace{0.18\linewidth}(a) Camera View 
        \hspace{0.195\linewidth}(b) BEV View
        \hspace{0.035\linewidth}(c) STMono3D
        \\
    \end{tabular}}
    \caption{\textbf{Depth-shift Illustration}. When inferring on the target domain, models can accurately locate the objects on the 2D image but predict totally wrong object depth with tremendous shifts. Such unreliable predictions for pseudo labels cannot improve but hurt the model performance in STMono3D. GAMS guarantees the geometry consistency and enables models predict correct object depth. Best view in color: prediction and ground truth are in \textcolor[rgb]{0.23,0.4,1}{blue} and \textcolor[rgb]{0.94,0.39,0.28}{orange}. Depth-shift is shown in \textcolor[rgb]{0,0.5,0}{green} arrows.} 
    \label{fig::depth-shift}
\end{figure}

However, 3D detectors developed on one specific dataset (\textit{i.e.} source domain) might suffer from tremendous performance degradation when generalizing to another dataset (\textit{i.e.} target domains) due to unavoidable domain-gaps arising from different types of sensors, weather conditions, and geographical locations. Especially, as shown in Fig.~\ref{fig::depth-shift}, the severe depth-shift caused by different imaging camera devices leads to totally failed localizations. As a result, a monocular detector trained on data collected in Singapore cities with NuScenes~\cite{caesar2020nusc} cameras \textbf{cannot} work well (\textit{i.e.,} average precision drops to zero) when evaluated on data from European cities captured by KITTI~\cite{geiger2012kitti} cameras. While collecting and training with more data from different domains could alleviate this problem, it is unfortunately infeasible, given diverse real-world scenarios and expensive annotation costs. Therefore, methods for effectively adapting a monocular 3D detector trained on a labeled source domain to a novel unlabeled target domain are highly demanded in practical applications. We call this task unsupervised domain adaptation (UDA) for monocular 3D object detection.

While intensive UDA studies~\cite{dubourvieux2021uda2d,long2015uda2d,hoffman2016uda2d,chen2018uda2d,saito2019uda2d,ge2020uda2d} on the 2D image setting are proposed, they mainly focus on handling lighting, color, and texture variations. However, in terms of the Mono3D, since detectors attend to estimate the spatial information of objects from monocular RGB images, the geometry alignment of domains is much more crucial. Moreover, for UDA on LiDAR-based 3D detection~\cite{yang2021st3d,yang2021st3d++,luo2021multialign,zhang2021srdan}, the fundamental differences in data structures and network architectures render these approaches not readily applicable to this problem. 

In this paper, we propose \textit{STMono3D}, for UDA on monocular 3D object detection. We first thoroughly investigate the depth-shift issue caused by the tight entanglement of models and camera parameters during the training stage. Models can accurately locate the objects on the 2D image but predict totally wrong object depth with tremendous shifts when inferring on the target domain. To alleviate this issue, we develop the \textit{geometry-aligned multi-scale} (GAMS) training strategy to guarantee the geometry consistency of domains and predict pixel-size depth to overcome the inevitable misalignment and ambiguity. Hence, models can provide effective predictions on the unlabeled target domain. Based upon this, we adopt the mean teacher~\cite{tarvainen2017mean} paradigm to facilitate the learning. The teacher model is essentially a temporal ensemble of student models, where parameters are updated by an exponential moving average window on student models of preceding iterations. It produces stable supervision for the student model without prior knowledge of the target domain. 


Moreover, we observe that the Mono3D teacher model suffers from extremely low confidence scores and numerous failed predictions on the target domain. To handle these issues, we adopt \textit{Quality-Aware Supervision} (QAS), \textit{Positive Focusing Training} (PFT), and \textit{Dynamic Threshold} (DT) strategies. Benefitting from the flexibility of the end-to-end mean teacher framework, we utilize the readability of each teacher-generated prediction to dynamically reweight the supervision loss of the student model, which takes instance-level qualities of pseudo labels into account, avoiding the low-quality samples interfering the training process. Since the backgrounds of domains are similar in the Mono3D UDA of the autonomous driving setting, we ignore the negative samples and only utilize positive pseudo labels to train the model. It avoids excessive FN pseudo labels at the beginning of the training process impairing the capability of the model to recognize objects. In synchronization with training, we utilize a dynamic threshold to adjust the filter score, which stabilizes the increase of pseudo labels.

To the best of our knowledge, this is the first study to explore effective UDA methods for Mono3D. Experimental results on various 3D object detection datasets KITTI~\cite{geiger2012kitti}, NuSenses~\cite{caesar2020nusc}, and Lyft~\cite{kesten2019lyft} demonstrate the effectiveness of our proposed methods, where the performance gaps between source only results and fully supervised oracle results are closed by a large margin. It is noteworthy that STMono3D even outperforms the oracle results under the NuScenes$\rightarrow$KITTI setting. Our codes will be available\footnote{https://github.com/zhyever/STMono3D}. 
\section{Related Work} 

\subsection{Monocular 3D Object Detection}
Mono3D has drawn increasing attention in recent years~\cite{chen20153d,xu2018multifusion,mousavian20173d,roddick2018orthographic,weng2019monocular,brazil2019m3drpn,wang2022detr3d,wang2021fcos3d,park2021dd3d,wang2022pgd}. Earlier work utilizes sub-networks to assist 3D detection. For instance, 3DOP~\cite{chen20153d} and MLFusion~\cite{xu2018multifusion} use a depth estimation network while Deep3DBox~\cite{mousavian20173d} uses a 2D object detector. Another line of research makes efforts to convert the RGB input to 3D representations like OFTNet~\cite{roddick2018orthographic} and Pseudo-Lidar~\cite{weng2019monocular}. While these methods have shown promising performance, they rely on the design and performance of sub-networks or dense depth labels. Hence, some methods propose to design the framework in an end-to-end manner like 2D detection. M3D-RPN~\cite{brazil2019m3drpn} implements a single-stage multi-class detector with a region proposal network and depth-aware convolution. SMOKE~\cite{liu2020smoke} proposes a neat framework to predict 3D objects without generating 2D proposals. DETR3D~\cite{wang2022detr3d} develop a DETR-like~\cite{carion2020end} bbox head, where 3D objects are predicted by independent queries in a set-to-set manner. DD3D~\cite{park2021dd3d} further investigates the influence of pre-trained monocular depth estimation network, in which they find depth estimation plays a crucial part in Mono3D. In this paper, we mainly conduct UDA experiments based on FCOS3D~\cite{wang2021fcos3d}, a neat and representative Mono3D paradigm that keeps the well-developed designs for 2D feature extraction and is adapted for this 3D task with only basic designs for specific 3D detection targets.

\subsection{Unsupervised Domain Adaptation}
UDA aims to generalize the model trained on a source domain to unlabeled target domains. So far, tremendous methods have been proposed for various computer vision tasks~\cite{dubourvieux2021uda2d,long2015uda2d,hoffman2016uda2d,chen2018uda2d,saito2019uda2d,ge2020uda2d,zou2018uda2d} (e.g., recognition, detection, segmentation). Some methods~\cite{mancini2018uda2d,sun2016uda2d,carlucci2017uda2d} employ the statistic-based metrics to model the differences between two domains. Other approaches~\cite{saito2017st2d,zou2018st2d,khodabandeh2019st2d} utilize the self-training strategy to generate pseudo labels for unlabeled target domains. Moreover, inspired by Generative Adversarial Networks (GANs)~\cite{goodfellow2014gan}, adversarial learning was employed to align feature distributions~\cite{tzeng2015uda2d,ganin2015uda2d,ganin2016uda2d}, which can be explained by minimizing the H-divergence~\cite{ben2010theory} or the Jensen-Shannon divergence~\cite{gulrajani2017improved} between two domains. \cite{li2018dabn2d,wang2019dabn2d} alleviated the domain shift on batch normalization layers by modulating the statistics in the BN layer before evaluation or specializing parameters of BN domain by domain. Most of these domain adaptation approaches are designed for the general 2D image recognition tasks, while direct adoption of these techniques for the large-scale monocular 3D object detection task may not work well due to the distinct characteristics of Mono3D, especially targes in 3D spatial coordination.

In terms of 3D object detection, \cite{zhang2021srdan,yang2021st3d,luo2021multialign} investigate UDA strategies for LIDAR-based detectors. SRDAN~\cite{zhang2021srdan} adopt adversarial losses to align the features and instances with similar scales between two domains. ST3D~\cite{yang2021st3d} and MLC-Net~\cite{luo2021multialign} develop self-training strategies with delicate designs, such as random object scaling, triplet memory bank, and multi-level alignment, for domain adaptation. Following the successful trend of UDA on LIDAR-based 3D object detection, we investigate effective self-training strategies for Mono3D. To the best of our knowledge, this is the first study to explore effective UDA methods for Mono3D.
\section{STMono3D} 

In this section, we first formulate the UDA task on Mono3D (Sec.~\ref{subsec::pd}), and present an overview of our framework (Sec.~\ref{subsec::fo}), followed by the Self-Teacher with Temporal Ensemble paradigm (Sec.~\ref{subsec::stte}). Then, we explain the details of the Geometry-Aligned Multi-Scale Training (GAMS, Sec.~\ref{subsec::gams}), the Quality-Aware Supervision (QAS, Sec.~\ref{subsec::qas}), and some other crucial training strategies consisting of Positive Focusing Training (PFT) and Dynamic Threshold (DT) (Sec.~\ref{subsec::cts}).

\subsection{Problem Definition}
\label{subsec::pd}
Under the unsupervised domain adaptation setting, we access to labeled images from the source domain $\mathcal{D}_S=\{x_s^i,y_s^i,K_s^i\}_{i=1}^{N_S}$, and unlabeled images from the target domain $\mathcal{D}_T=\{x_t^i, K_t^i\}_{i=1}^{N_T}$, where $N_s$ and $N_t$ are the number of samples from the source and target domains, respectively. Each 2D image $x^i$ is paired with a camera parameter $K^i$ that projects points in 3D space to 2D image plane while $y_s^i$ denotes the label of the corresponding training sample in the specific camera coordinate from the source domain. Label $y$ is in the form of object class $k$, location $(c_x, c_y, c_z)$, size in each dimension $(d_x, d_y, d_z)$, and orientation $\theta$. We aim to train models with $\{\mathcal{D}_S, \mathcal{D}_T\}$ and avoid performance degradation when evaluating on the target domain.

\begin{figure}[t]
    \includegraphics[width=1\linewidth]{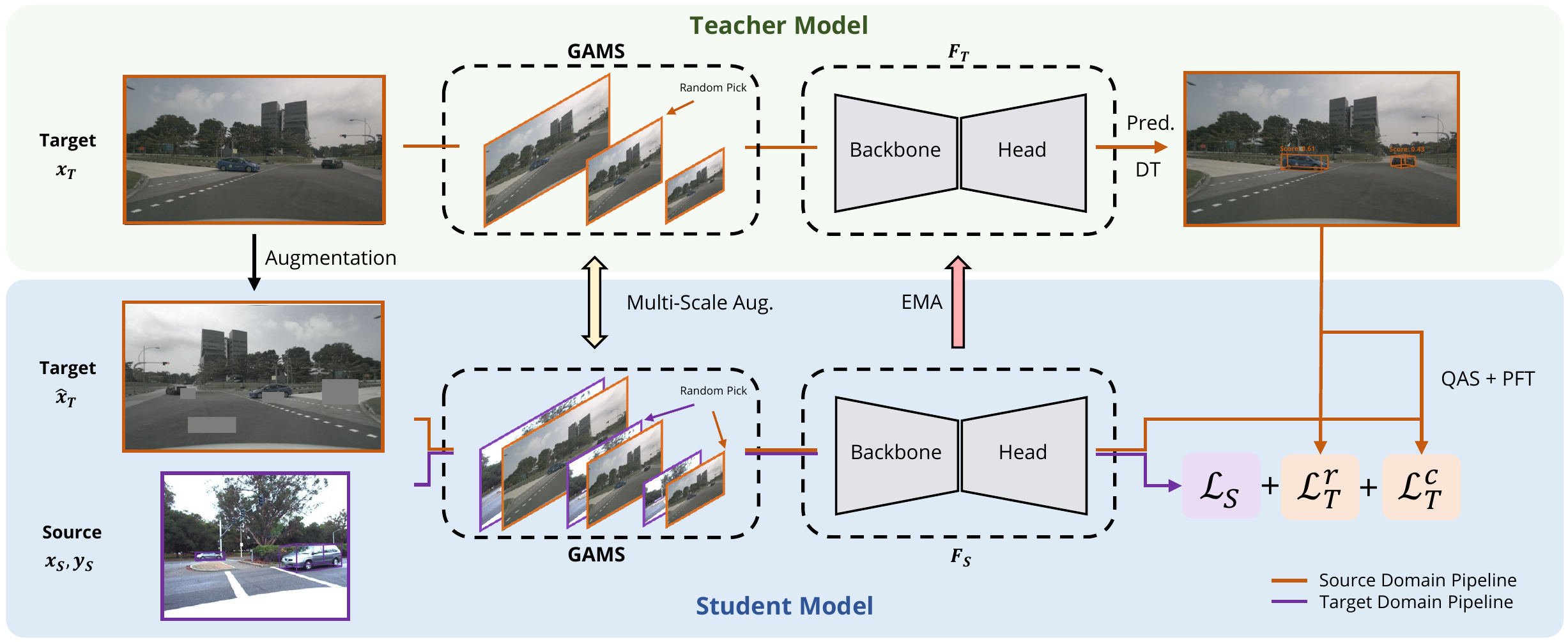}
    \caption{Framework overview. STMono3D leverages the mean-teacher~\cite{tarvainen2017mean} paradigm where the teacher model is the exponential moving average of the student model and updated at each iteration. We design the GAMS (Sec.~\ref{subsec::gams}) to alleviate the severe depth-shift in cross domain inference and ensure the availability of pseudo labels predicted by the teacher model. QAS (Sec.~\ref{subsec::qas}) is a simple \textit{soft-teacher} approach which leverages richer information from the teacher model to reweight losses and provide quality-aware supervision on the student model. PFT and DT are another two crucial training strategies presented in Sec. \ref{subsec::cts}.}
    \label{fig::framework}
\end{figure}

\subsection{Framework Overview}
\label{subsec::fo}
We illustrate our STMono3D in Fig.~\ref{fig::framework}. The labeled source domain data $\{x_S, y_S\}$ is utilized for supervised training of the student model $F_S$ with a loss $\mathcal{L}_S$. In terms of the unlabeled target domain data $x_T$, we first perturb it by applying a strong random augmentation to obtain $\hat{x}_T$. Before passing to the models, both the target and source domain input are further augmented by the GAMS strategy in Sec.~\ref{subsec::gams}, where images and camera intrinsic parameters are cautiously aligned via simultaneously rescaling. Subsequently, the original and perturbed images are sent to the teacher and student model, respectively, where the teacher model generates intuitively reasonable pseudo labels $\hat{y}_T$ and supervises the student model via loss $\mathcal{L}_T$ on the target domain:
\begin{equation}
    \mathcal{L}_T = \mathcal{L}_{T}^{r} + \mathcal{L}_{T}^{c},
\end{equation}
where $\mathcal{L}_{T}^{r}$ and $\mathcal{L}_{T}^{c}$ are the regression loss and classification loss, respectively. Here, we adopt the QAS strategy in Sec.~\ref{subsec::qas} to further leverage richer information from the teacher model by instance-wise reweighting the loss $\mathcal{L}_T$. In each iteration, the student model is updated through gradient descent with the total loss $\mathcal{L}$, which is a linear combination of $\mathcal{L}_S$ and $\mathcal{L}_T$:
\begin{equation}
\label{eq::total-loss}
\mathcal{L} = \lambda\mathcal{L}_S + \mathcal{L}_T,
\end{equation}
where $\lambda$ is the weight coefficient. Then, the teacher model parameters are updated by the corresponding parameters of the student model, where we introduce the details in Sec.~\ref{subsec::stte}. Moreover, we observe that the teacher model suffers from numerous FN and FP pseudo labels on the target domain. To handle this issue, we utilize the PFT and DT strategies illustrated in Sec.~\ref{subsec::cts}.




\subsection{Self-Teacher with Temporal Ensemble}
\label{subsec::stte}
Following the successful trend of the mean teacher paradigm~\cite{tarvainen2017mean} in the semi-supervised learning, we adapt it to our Mono3D UDA task as illustrated in Fig.~\ref{fig::framework}. The teacher model $F_T$ and the student model $F_S$ share the same network architecture but have different parameters $\theta_T$ and $\theta_S$, respectively. During the training, the parameters of the teacher model are updated via taking the exponential moving average (EMA) of the student parameters:
\begin{equation}
\label{eq::mm}
    \theta_T = m\theta_T + (1-m)\theta_S,
\end{equation}
where $m$ is the momentum that is commonly set close to 1, \textit{e.g.,} 0.999 in our experiments. Moreover, the input of the student model is perturbed by a strong augmentation, which ensures that the pseudo labels generated by the teacher model are more accurate than the student model predictions, thus providing available optimization directions for the parameter updating. In addition, the strong augmentation can also improve the model generalization to handle the different domain inputs. Hence, by supervising the student model with pseudo targets $\hat{y}_T$ generated by the teacher model (\textit{i.e.,} forcing the consistency between predictions of the student and the teacher model), the student can learn domain-invariant representations to adapt to the unlabeled target domain. Fig.~\ref{fig::teacher} shows that the teacher model can provide effective supervision to the student model and Tab.~\ref{tab::ablation_teacher},~\ref{tab::ablation_ema} demonstrate the effectiveness of the mean teacher paradigm.

\subsection{Geometry-Aligned Multi-Scale Training}
\label{subsec::gams}
\subsubsection{Observation.} As shown in Fig.~\ref{fig::depth-shift}, depth-shift drastically harms the quality of pseudo labels on the target domain. It is mainly caused by the domain-specific geometry correspondences between 3D objects and images (\textit{i.e.,} camera imaging process). For instance, since the pixel size (defined in Eq.~\ref{eq::pixel-size}) of the KITTI dataset is larger than the NuScenes dataset, objects in images captured by KITTI cameras are smaller than NuScenes ones. While the model can predict accurate 2D locations on image planes, it tends to estimate relatively more significant object depth based on the depth cue that far objects tend to be smaller in perspective view. We call the phenomenon depth-shift: models localize accurate 2D location but predict depth with tremendous shifts on the target domain. To mitigate it, we propose a straight-forward yet effective augmentation strategy, \textit{i.e., geometry-aligned multi-scale} training, fully leveraging the geometry consistency in the imaging process.

\subsubsection{Method.} Given the source input $\{x_S, y_S, K_S\}$ and the target input $\{x_T, K_T\}$, a naive geometry-aligned strategy is to rescale camera parameters to the same constant values and resize images correspondingly:
\begin{equation}
    \mathbf{K} =
    \left[\begin{array}{ccc} r_x ~~& r_y ~~& 1\end{array} \right]
    \left[\begin{array}{ccc} f_x ~~& 0 ~~& p_x\\
                             0 ~~& f_y ~~& p_y\\
                             0 ~~& 0 ~~& 1\end{array} \right]
\end{equation}
where $r_x$ and $r_y$ are resize rates, $f$ and $p$ are focal length and optical center, $x$ and $y$ indicate image coordinate axises, respectively. However, since the $f/p$ cannot be changed by resizing, it is impracticable to strictly align the geometry correspondences of 3D objects and images between different domains via convenient transformations. The inevitable discrepancy and ambiguity lead to a failure on UDA.

To solve the issue, motivated by DD3D~\cite{park2021dd3d}, we propose to predict the \textit{pixel-size depth} $d_p$ instead of the \textit{metric depth} $d_g$:
\begin{equation}
    d_p = \frac{s}{c}\cdot d_g,
\end{equation}
\begin{equation}
\label{eq::pixel-size}
    s = \sqrt{\frac{1}{f_x^2} + \frac{1}{f_y^2}},
\end{equation}
where $s$ and $c$ are the pixel size and a constant, $d_p$ is the model prediction and is scaled to the final result $d_g$. Hence, while there are inevitable discrepancies between aligned geometry correspondences of two domains, the model can infer the depth from the pixel size and be more robust to the various imaging process. Moreover, we further rescale camera parameters into a multi-scale range, instead of the same constant values, and resize images correspondingly to enhance the dynamic of model. During the training, we keep ground-truth 3D bounding boxes $y_S$ and pseudo labels $\hat{y}_T$ unchanged, but modify camera parameters and image resolutions simultaneously.

\subsection{Quality-Aware Supervision}
\label{subsec::qas}
\subsubsection{Observation.} 
\begin{figure}[t]
    \centering
    \footnotesize
    \scalebox{0.98}{%
    \begin{tabular}{ccc}
        \includegraphics[width=0.3\linewidth]{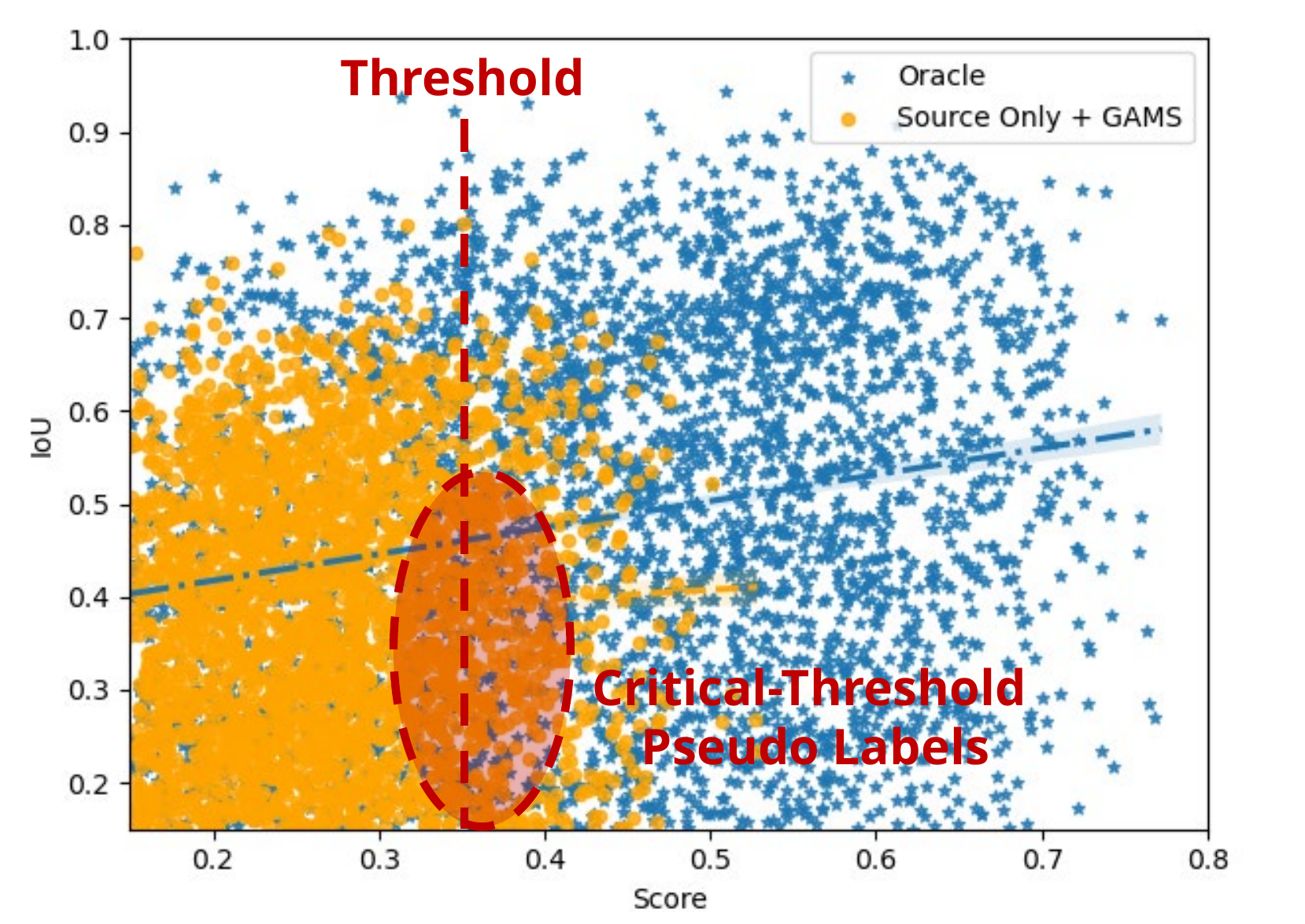}&
        \includegraphics[width=0.3\linewidth]{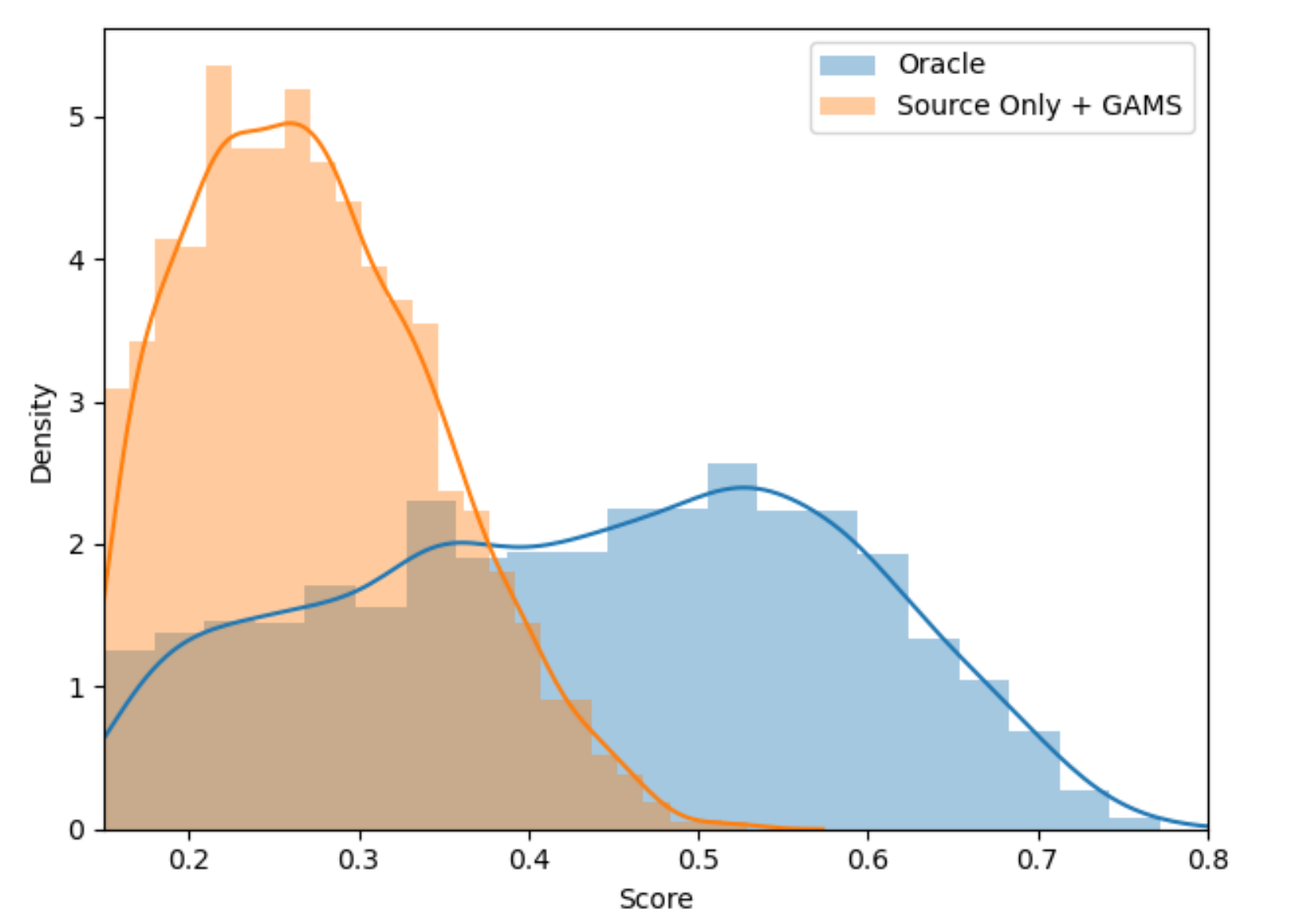}&
        \includegraphics[width=0.3\linewidth]{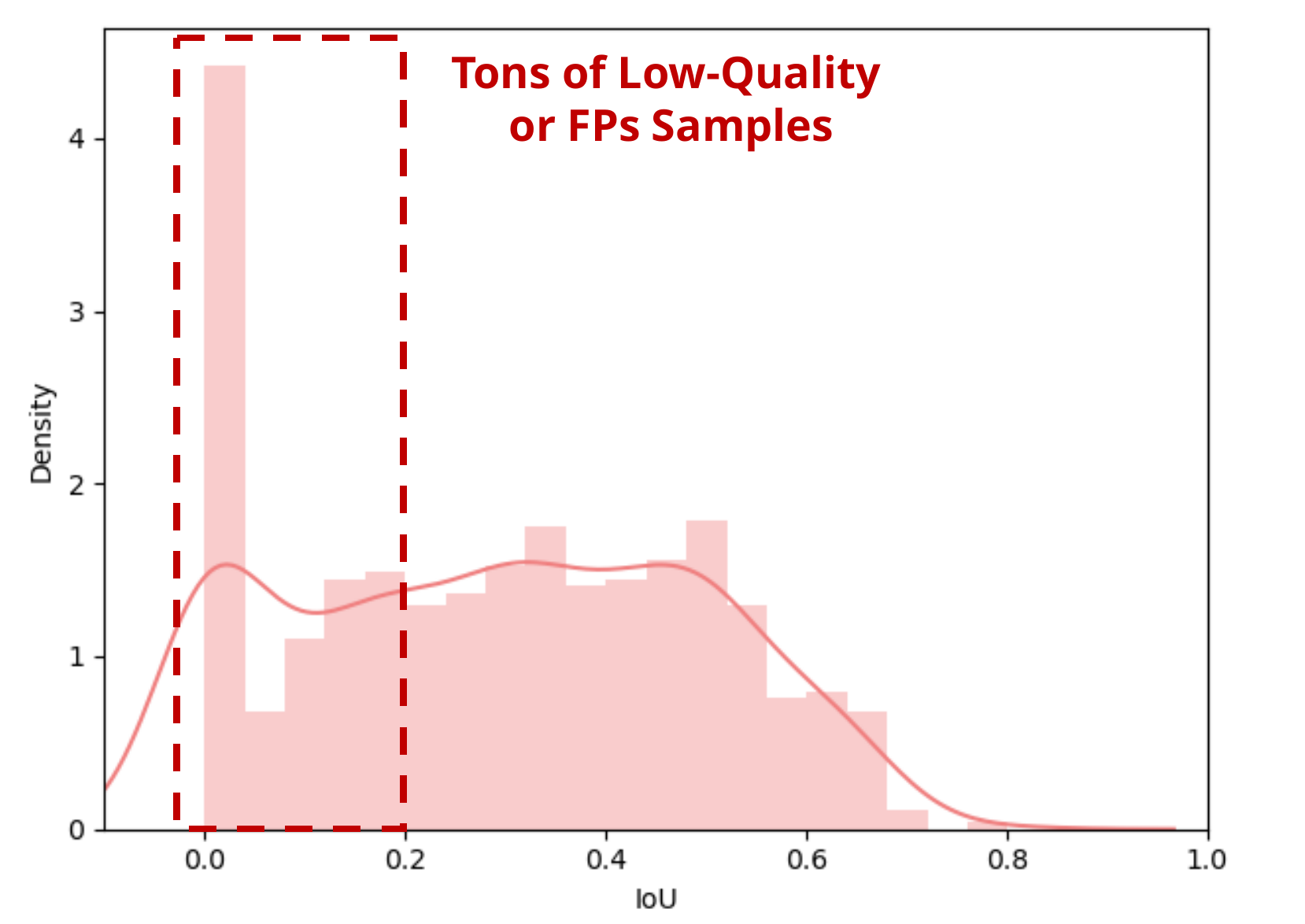}\\
        \scriptsize{(a)} & \scriptsize{(b)} & \scriptsize{(c)} \\
    \end{tabular}}
    \vspace{-0.15cm}
    \caption{(a) Correlation between confidence value and box IoU with ground-truth. (b) Distribution of confidence scores. The teacher suffers from low scores on the target domain. (c) Distribution of IoU between ground-truth and pseudo labels near the threshold (0.35-0.4). We highlight the existence of numerous low-quality and FP samples in these pseudo labels.}
    \label{fig::score-iou}
\end{figure}

The cross-domain performance of the detector highly depends on the quality of pseudo labels. In practice, we have to utilize a higher threshold on the foreground score to filter out most false positive (FP) box candidates with low confidence. However, unlike the teacher model that can detect objects with high confidence in the semi-supervised 2D detection or UDA of LiDAR-based 3D detector (\textit{e.g.,} the threshold is set to 90\% and 70\% in \cite{xu2021soft} and \cite{yang2021st3d}, respectively), we find the Mono3D cross-domain teacher \textbf{suffers from a much lower confidence} as shown in Fig.~\ref{fig::score-iou}, which is another unique phenomenon in Mono3D UDA caused by the much worse oracle monocular 3D detection performance than 2D detection and LiDAR-based 3D detection, which indicates that though the prediction confidence surpasses the threshold, we cannot ensure the sample quality, especially for the ones near the threshold. To alleviate the impact, we propose the \textit{quality-aware supervision} (QAS) to leverage richer information from the teacher and take instance-level quality into account.

\subsubsection{Method.} Thanks to the flexibility of the end-to-end mean teacher framework, we assess the reliability of each teacher-generated bbox to be a real foreground, which is then used to weight the foreground classification loss of the student model. Given the foreground bounding box set $\{b_i^{fg}\}_{i=1}^{N^{fg}}$, the classification loss of the unlabeled images on the target domain is defined as:
\begin{equation}
    \mathcal{L}_T^c = \frac{\mu}{N^{fg}}\sum\limits_{i=1}^{N^{fg}}w_i\cdot l_{cls}(b_i^{fg}, \mathcal{G}_{cls}),
\end{equation}
where $\mathcal{G}_{cls}$ denotes the set of pseudo class labels, $l_{cls}$ is the box classification loss, $w_i$ is the confidence score for $i^{th}$ foreground pseudo boxes, $N^{fg}$ is the number of foreground pseudo box, and $\mu$ is a constant hyperparameter.

The QAS resembles a \textit{simple positive mining} strategy, which is intuitively reasonable that there should be more severe punishment for pseudo labels with higher confidence. Moreover, compared with semi-supervised and supervised tasks that focus on simple/hard negative samples~\cite{xu2021soft,chen2021semiseg}, it is more critical for UDA Mono3D models to prevent harmful influence caused by low-quality pseudo labels near the threshold. Such an instance-level weighting strategy balances the loss terms based on foreground confidence scores and significantly improves the effectiveness of STMono3D.
 
\subsection{Crucial Training Strategies}
\label{subsec::cts}
\subsubsection{Positive Focusing Training.}
Since the whole STMono3D is trained in an end-to-end manner, the teacher model can hardly detect objects with confident scores higher than the threshold at the start of the training. Tons of FN pseudo samples impair the capability of the model to recognize objects. Because backgrounds of different domains are similar with negligible domain gaps in Mono3D UDA (\textit{e.g.,} street, sky, and house), we propose the \textit{positive focusing training} strategy. As for the $\mathcal{L}_T^c$, we discard negative background pseudo labels and only utilize the positive samples to supervise the student model, which ensures that the model does not crash to overfit on the FN pseudo labels during the training stage. 

\subsubsection{Dynamic Threshold.}
In practice, we find that the mean confidence score of pseudo labels gradually increases in synchronization within training duration. Increasing false positive (FP) samples appear in the middle and late stages of training, which harshly hurts the model performance. While the QAS strategy proposed in Sec.~\ref{subsec::qas} can reduce the negative impact of low-quality pseudo labels, the completely wrong predictions still introduce inevitable noise to the training process. To alleviate the issue, we propose a simple \textit{progressively increasing threshold} strategy to dynamic change the threshold $\tau$ as:
\begin{equation}
\tau=\left\{
\begin{array}{lcl}
\alpha,    &      & {iter < n_1},\\
\alpha + k\cdot (iter- n_1),   &  ~~~~ & {n_1 \leqslant iter < n_2}, \\
\alpha + k\cdot (n_2 - n_1),   &   & {iter \geqslant n_2},
\end{array} \right.
\end{equation}
where $\alpha$ is the base threshold that is set to 0.35 based on the statistics in Fig.~\ref{fig::score-iou}(a) in our experiments, $k$ is the slope of increasing threshold, $iter$ is the iteration of training stage. The threshold is fixed to a minimum during the first $n$ warmup steps as the teacher model can hardly detect objects with confident scores higher than the base threshold. It then linearly increases after the teacher model predicts pseudo labels with FP samples to avoid the model being blemished by increasing failure predictions. Finally, we find that the increasing of average scores tends to a saturation. Therefore, the threshold is fixed at the end of the training stage to guarantee the number of pseudo labels.

\begin{table}[b]
    \centering
    \caption{Dataset Overview. We focus on their properties related to frontal-view cameras and 3D object detection. The dataset size refers to the number of images used in training stage. For Waymo and NuScenes, we subsample the data. See text for details.}
    \scalebox{0.9}{%
    \begin{tabular}{cccccccc}
        \hline
        ~~Dataset~~ & ~~~Size~~~ & ~~~Anno.~~~ & ~~~Loc.~~~ & ~~~~Shape~~~~ & ~~~~FOV~~~~ & ~Objects~ & ~~Night~~\\
        \hline
        KITTI~\cite{geiger2012kitti} & 3712 & 17297 & EUR. & (375,1242) & (29$^{\circ}$,81$^{\circ}$) & 8 & No\\
        NuScenes~\cite{caesar2020nusc} & 27522 & 252427 & SG.,EUR. & (900,1600) & (39$^{\circ}$,65$^{\circ}$) & 23 & Yes\\
        Lyft~\cite{kesten2019lyft} & 21623 & 139793 & SG.,EUR. & (1024,1224) & (60$^{\circ}$,70$^{\circ}$) & 9 & No\\
        \hline
    \end{tabular}
    }
\label{tab::data_info}
\vspace{-0.2cm}
\end{table}

\section{Experiments}
\label{sec:experiments}

\subsection{Experimental Setup}
\subsubsection{Datasets.}
We conduct experiments on three widely used autonomous driving datasets: KITTI~\cite{geiger2012kitti}, NuSenses~\cite{caesar2020nusc}, and Lyft~\cite{kesten2019lyft}. Two aspects are lying in our experiments: Cross domains with different cameras (existing in all the source-target pairs) and adaptation from label rich domains to insufficient domains (\textit{i.e.,} NuSenses$\rightarrow$KITTI). We summarize the dataset information in detail in Tab.~\ref{tab::data_info}, and present more visualization comparisons in the \textit{supplementary material}.

\subsubsection{Comparison Methods.}
In our experiments, we compare STMono3D with three methods: ($i$) \textbf{Source Only} indicates directly evaluating the source domain trained model on the target domain. ($ii$) \textbf{Oracle} indicates the fully supervised model trained on the target domain. ($iii$) \textbf{Naive ST (with GAMS)} is the basic self-training method. We first train a model (with GAMS) on the source domain, then generate pseudo labels for the target domain, and finally fine-tuning the trained model on the target domain.

\subsubsection{Evaluation Metric.}
We adopt the KITTI evaluation metric for evaluating our methods in NuSenses$\rightarrow$KITTI and Lyft$\rightarrow$KITTI and the NuScenes metric for Lyft$\rightarrow$NuSenses. We focus on the commonly used car category in our experiments. For Lyft$\rightarrow$NuSenses, we evaluate models on ring view, which is more useful in real-world applications. For KITTI, We report the average precision (AP) where the IoU thresholds are 0.5 for both the bird's eye view (BEV) IoUs and 3D IoUs. For NuScenes, since the attribute labels are different from the source domain (\textit{i.e.,} Lyft), we discard the average attribute error (mAAE) and report the average trans error (mATE), scale error (mASE), orient error (mAOE), and average precision (mAP). Following~\cite{yang2021st3d}, we report the closed performance gap between Source Only to Oracle.

\subsubsection{Implementation Details.}
We validate our proposed STMono3D on detection backbone FCOS3D~\cite{wang2021fcos3d}. Since there is no modification to the model, our method can be adapted to other Mono3D backbones as well. We implement STMono3D based on the popular 3D object detection codebase mmDetection3D~\cite{mmdet3d2020}. We utilize SGD~\cite{kingma2014adam} optimizer. Gradient clip and warm-up policy are exploited with the learning rate $2\times10^{-2}$, the number of warm-up iterations 500, warm-up ratio 0.33, and batch size 32 on 8 Tesla V100s. The loss weight $\lambda$ of different domains in Eq.~\ref{eq::total-loss} is set to 1. We apply a momentum $m$ of 0.999 in Eq.~\ref{eq::mm} following most of mean teacher paradigms~\cite{luo2021multialign,xu2021soft}. As for the strong augmentation, we adopt the widely used image data augmentation, including random flipping, random erase, random toning, \textit{etc}. We subsample $\frac{1}{4}$ dataset during the training stage of NuScenes and Lyft dataset for simplicity. Notably, unlike the mean teacher paradigm or the self-training strategy used in UDA of LiDAR-based 3D detector~\cite{luo2021multialign,yang2021st3d}, our STMono3D is trained in a \textit{totally end-to-end} manner.

\subsection{Main Results}
As shown in Tab.~\ref{tab::overall}, we compare the performance of our STMono3D with Source Only and Oracle. Our method outperforms the Source Only baseline on all evaluated UDA settings. Caused by the domain gap, the Source Only model cannot detect 3D objects where the mAP almost drops to 0\%. Otherwise, STMono3D improves the performance on NuScenes$\rightarrow$KITTI and Lyft$\rightarrow$KITTI tasks by a large margin that around 110\%/67\% performance gap of $AP_{3D}$ are closed. Notably, the $AP_{BEV}$ and $AP_{3D}$ of $AP_{11}, IoU\geqslant0.5$ of STMono3D surpass the Oracle results, which indicates the effectiveness of our method. Furthermore, when transferring Lyft models to other domains that have full ring view annotations for evaluation (\textit{i.e.}, Lyft$\rightarrow$NuScenes), our STMono3D also attains a considerable performance gain which closes the Oracle and Source Only performance gap by up to 66\% on $AP_{3D}$. These encouraging results validate that our method can effectively adapt 3D object detectors to the target domain.

\begin{table}[t]
    \centering
    \caption{Performance of STMono3D on three source-target pairs. We report $AP$ of the car category at $\text{IoU} = 0.5$ as well as the domain gap closed by STMono3D. In Nus$\rightarrow$KITTI, STMono3D achieves a even better results on $AP_{11}$ compared with the Oracle model, which demonstrates the effectiveness of our proposed method.}
    \scalebox{0.74}{%
    \begin{tabular}{|c|ccc|ccc|ccc|ccc|}
        \hline
        \textbf{Nus$\rightarrow$K} & \multicolumn{6}{c|}{$AP_{11}$} & \multicolumn{6}{c|}{$AP_{40}$} \\\hline
        \multirow{2}{*}{Method}&\multicolumn{3}{c|}{$AP_{BEV}$ $\text{IoU}\geqslant0.5$} & \multicolumn{3}{c|}{$AP_{3D}$ $\text{IoU}\geqslant0.5$} & \multicolumn{3}{c|}{$AP_{BEV}$ $\text{IoU}\geqslant0.5$} & \multicolumn{3}{c|}{$AP_{3D}$ $\text{IoU}\geqslant0.5$} \\
        & Easy     & Mod.     & Hard   & ~Easy~     & Mod.     & Hard & ~Easy~     & ~~Mod.~     & ~~Hard~  & Easy     & Mod.     & Hard \\ 
        \hline
        Source Only & 0 & 0 & 0 & 0 & 0 & 0 & 0 & 0 & 0 & 0 & 0 & 0 \\
        Oracle & 33.46 & 23.62 & 22.18 & 29.01 & 19.88 & 17.17 & 33.70 & 23.22 & 20.68 & 28.33 & 18.97 & 16.57\\
        STMono3D & 35.63 & 27.37 & 23.95 & 28.65 & 21.89 & 19.55 
        & 31.85 & 22.82 & 19.30 & 24.00 & 16.85 & 13.66 \\
        \hline
        \cellcolor{LightCyan}Closed Gap & \cellcolor{LightCyan}106.5\% & \cellcolor{LightCyan}115.8\% & \cellcolor{LightCyan}107.9\% & \cellcolor{LightCyan}98.7\% & \cellcolor{LightCyan}110.1\% & \cellcolor{LightCyan}113.8\% & \cellcolor{LightCyan}~94.5\% & \cellcolor{LightCyan}98.2\% & \cellcolor{LightCyan}93.3\% & \cellcolor{LightCyan}84.7\% & \cellcolor{LightCyan}88.8\% & \cellcolor{LightCyan}82.4\%\\
        \hline
        \hline
        \textbf{L$\rightarrow$K} & \multicolumn{6}{c|}{$AP_{11}$} &\multicolumn{2}{c|}{\textbf{L$\rightarrow$Nus}} & \multicolumn{4}{c|}{~~~~~~~~~~~~~~~~Metrics~~~~~~~~~~~~~~~~~~}\\
        \hline
        \multirow{2}{*}{Method}&\multicolumn{3}{c|}{$AP_{BEV}$ $\text{IoU}\geqslant0.5$} & \multicolumn{3}{c|}{$AP_{3D}$ $\text{IoU}\geqslant0.5$} & \multicolumn{2}{c|}{\multirow{2}{*}{Method}} & \multirow{2}{*}{~~AP~~} & \multirow{2}{*}{~~ATE~~} & \multirow{2}{*}{~~ASE~~} & \multirow{2}{*}{~~AOE~~} \\
        & Easy     & Mod.     & Hard   & Easy     & Mod. & Hard & \multicolumn{2}{c|}{} &&&& \\ 
        \hline
        Source Only & 0 & 0 & 0 & 0 &  0 & 0 & \multicolumn{2}{c|}{Source Only} & 2.40 & 1.302 & 0.190 & 0.802 \\
        Oracle & 33.46 & 23.62 & 22.18 & 29.01 & 19.88 & 17.17 & \multicolumn{2}{c|}{Oracle} & 28.2 & 0.798 & 0.160 & 0.209 \\ 
        STMono3D & 26.46 & 20.71 & 17.66 & 18.14 & 13.32 & 11.83 & \multicolumn{2}{c|}{STMono3D} & 21.3 & 0.911 & 0.170 & 0.355\\
        \hline
        \cellcolor{LightCyan}Closed Gap & \cellcolor{LightCyan}79.0\% & \cellcolor{LightCyan}87.6\% & \cellcolor{LightCyan}79.6\% & \cellcolor{LightCyan}62.5\% & \cellcolor{LightCyan}67.0\% & \cellcolor{LightCyan}68.8\% & \multicolumn{2}{c|}{\cellcolor{LightCyan}Closed Gap} & \cellcolor{LightCyan}73.2\% & \cellcolor{LightCyan}77.5\% & \cellcolor{LightCyan}66.7\% & \cellcolor{LightCyan}82.9\%\\
        \hline
    \end{tabular}
    }
\label{tab::overall}
\vspace{-0.2cm}
\end{table}

\begin{table}[t]
    \centering
    \caption{Ablation study of the geometry-aligned multi-scale training.}
    \scalebox{0.8}{%
    \begin{tabular}{|c|ccc|ccc|ccc|ccc|}
        \hline
        \textbf{Nus$\rightarrow$K} & \multicolumn{6}{c|}{$AP_{11}$} & \multicolumn{6}{c|}{$AP_{40}$} \\\hline
        \multirow{2}{*}{GAMS} &\multicolumn{3}{c|}{$AP_{BEV}$ $\text{IoU}\geqslant0.5$} & \multicolumn{3}{c|}{$AP_{3D}$ $\text{IoU}\geqslant0.5$} & \multicolumn{3}{c|}{$AP_{BEV}$ $\text{IoU}\geqslant0.5$} & \multicolumn{3}{c|}{$AP_{3D}$ $\text{IoU}\geqslant0.5$} \\
        & Easy & Mod. & Hard & Easy & Mod. & Hard & Easy & Mod. & Hard & Easy & Mod. & Hard \\ 
        \hline
        & 0 & 0& 0 & 0 & 0 & 0 & 0 & 0 & 0 & 0 & 0 & 0 \\
        \cellcolor{LightCyan}$\surd$& \cellcolor{LightCyan}35.63 & \cellcolor{LightCyan}27.37 & \cellcolor{LightCyan}23.95 & \cellcolor{LightCyan}28.65 & \cellcolor{LightCyan}21.89 & \cellcolor{LightCyan}19.55 & \cellcolor{LightCyan}31.85 & \cellcolor{LightCyan}22.82 & \cellcolor{LightCyan}19.30 & \cellcolor{LightCyan}24.00 & \cellcolor{LightCyan}16.85 & \cellcolor{LightCyan}13.66 \\
        \hline
    \end{tabular}
    }
\label{tab::ablation_GAMS}
\vspace{-0.2cm}
\end{table}

\subsection{Ablation Studies and Analysis}
In this section, we conduct extensive ablation experiments to investigate the individual components of our STMono3D. All experiments are conducted on the task of NuScenes$\rightarrow$KITTI.
\subsubsection{Effective of Geometry-Aligned Multi-Scale Training.}
We study the effects of GAMS in the mean teacher paradigm of STMono3D and the Naive ST pipeline. Tab.~\ref{tab::ablation_GAMS} first reports the experimental results when GAMS is disabled. Caused by the depth-shift analyzed in Sec.~\ref{subsec::gams}, the teacher model generates incorrect pseudo labels on the target domain, thus leading to a severe drop in model performance. Furthermore, as shown in Tab.~\ref{tab::ablation_teacher}, GAMS is crucial for effective Naive ST as well. It is reasonable that GAMS supports the model trained on the source domain to generate valid pseudo labels on the target domain, making the fine-tuning stage helpful for the model performance. We present pseudo labels predicted by the teacher model of STMono3D in Fig.~\ref{fig::depth-shift}, which shows that the depth-shift is well alleviated. All the results highlight the importance of GAMS for effective Mono3D UDA.

\begin{table}[t]
    \centering
    \caption{Comparison of different self-training paradigms.}
    \scalebox{0.8}{%
    \begin{tabular}{|c|ccc|ccc|cccc|}
        \hline
        \textbf{Nus$\rightarrow$K} & \multicolumn{6}{c|}{KITTI $AP_{40}$} & \multicolumn{4}{c|}{Nus Metrics} \\\hline
        \multirow{2}{*}{Method} & \multicolumn{3}{c|}{$AP_{BEV}$ $\text{IoU}\geqslant0.5$} &  \multicolumn{3}{c|}{$AP_{3D}$ $\text{IoU}\geqslant0.5$} & \multirow{2}{*}{~AP~} & \multirow{2}{*}{~ATE~} & \multirow{2}{*}{~ASE~} & \multirow{2}{*}{~AOE~}\\
        & Easy & Mod. & Hard & Easy & Mod. & Hard  & & & & \\ 
        \hline
        Naive ST  & 0 & 0 & 0 & 0 & 0 & 0 & 0 & 0 & 0 & 0 \\
        Naive ST with GAMS & 9.05 & 9.08 & 8.82 & 3.72 & 3.69 & 3.58 & 14.0 & 0.906 & 0.164 & 0.264 \\
        \cellcolor{LightCyan}STMono3D & \cellcolor{LightCyan}35.63 & \cellcolor{LightCyan}27.37 & \cellcolor{LightCyan}23.95 & \cellcolor{LightCyan}28.65 & \cellcolor{LightCyan}21.89 & \cellcolor{LightCyan}19.55 & \cellcolor{LightCyan}36.5 & \cellcolor{LightCyan}0.731 & \cellcolor{LightCyan}0.160 & \cellcolor{LightCyan}0.167 \\
        \hline
    \end{tabular}
    }
\label{tab::ablation_teacher}
\vspace{-0.2cm}
\end{table}

\begin{table}[t]
    \centering
    \caption{Ablation study of the exponential moving average strategy.}
    \scalebox{0.8}{%
    \begin{tabular}{|c|ccc|ccc|ccc|ccc|}
        \hline
        \textbf{Nus$\rightarrow$K} & \multicolumn{6}{c|}{$AP_{11}$} & \multicolumn{6}{c|}{$AP_{40}$} \\\hline
        \multirow{2}{*}{EMA} &\multicolumn{3}{c|}{$AP_{BEV}$ $\text{IoU}\geqslant0.5$} & \multicolumn{3}{c|}{$AP_{3D}$ $\text{IoU}\geqslant0.5$} & \multicolumn{3}{c|}{$AP_{BEV}$ $\text{IoU}\geqslant0.5$} & \multicolumn{3}{c|}{$AP_{3D}$ $\text{IoU}\geqslant0.5$} \\
        & Easy & Mod. & Hard & Easy & Mod. & Hard & Easy & Mod. & Hard & Easy & Mod. & Hard \\ 
        \hline
        & 2.55 & 2.41 & 2.38 & 0.82 & 0.82 & 0.82 & 0.45 & 0.31 & 0.25 & 0.06 & 0.03 & 0.02 \\
        \cellcolor{LightCyan}$\surd$& \cellcolor{LightCyan}35.63 & \cellcolor{LightCyan}27.37 & \cellcolor{LightCyan}23.95 & \cellcolor{LightCyan}28.65 & \cellcolor{LightCyan}21.89 & \cellcolor{LightCyan}19.55 & \cellcolor{LightCyan}31.85 & \cellcolor{LightCyan}22.82 & \cellcolor{LightCyan}19.30 & \cellcolor{LightCyan}24.00 & \cellcolor{LightCyan}16.85 & \cellcolor{LightCyan}13.66 \\
        \hline
    \end{tabular}
    }
\label{tab::ablation_ema}
\vspace{-0.2cm}
\end{table}

\begin{table}[t]
    \centering
    \caption{Ablation study of QAS on different loss terms.}
    \scalebox{0.8}{%
    \begin{tabular}{|cc|ccc|ccc|ccc|ccc|}
        \hline
        \multicolumn{2}{|c|}{\textbf{Nus$\rightarrow$K}} & \multicolumn{6}{c|}{$AP_{11}$} & \multicolumn{6}{c|}{$AP_{40}$} \\\hline
        \multirow{2}{*}{$L_{T}^{reg}$} & \multirow{2}{*}{$L_{T}^{cls}$} &\multicolumn{3}{c|}{$AP_{BEV}$ $\text{IoU}\geqslant0.5$} & \multicolumn{3}{c|}{$AP_{3D}$ $\text{IoU}\geqslant0.5$} & \multicolumn{3}{c|}{$AP_{BEV}$ $\text{IoU}\geqslant0.5$} & \multicolumn{3}{c|}{$AP_{3D}$ $\text{IoU}\geqslant0.5$} \\
        & & Easy & Mod. & Hard & Easy & Mod. & Hard & Easy & Mod. & Hard & Easy & Mod. & Hard \\ 
        \hline
        & & 26.33 & 21.92 & 19.57 & 21.17 & 18.14 & 16.46 & 21.66& 16.64& 14.03& 15.55& 12.06& 9.88 \\
        $\surd$& & 21.50 & 17.57 & 15.35 & 16.57 & 13.80 & 11.34 & 20.47 & 15.77 & 13.12 & 15.32 & 11.69 & 9.35 \\
        \cellcolor{LightCyan}& \cellcolor{LightCyan}$\surd$& \cellcolor{LightCyan}35.63 & \cellcolor{LightCyan}27.37 & \cellcolor{LightCyan}23.95 & \cellcolor{LightCyan}28.65 & \cellcolor{LightCyan}21.89 & \cellcolor{LightCyan}19.55 & \cellcolor{LightCyan}31.85 & \cellcolor{LightCyan}22.82 & \cellcolor{LightCyan}19.30 & \cellcolor{LightCyan}24.00 & \cellcolor{LightCyan}16.85 & \cellcolor{LightCyan}13.66 \\
        $\surd$& $\surd$& 21.74 & 19.56 & 17.22 & 18.09 & 15.67 & 14.71 
        & 16.01 & 13.26 & 11.15 & 10.89 & 9.22 & 7.49 \\
        \hline
    \end{tabular}
    }
\label{tab::ablation_QAS}
\vspace{-0.2cm}
\end{table}

\begin{figure}[t]
    \centering
    \footnotesize
    \setlength{\tabcolsep}{10pt}
    \scalebox{0.95}{%
    \begin{tabular}{cc}
        \includegraphics[width=0.45\linewidth]{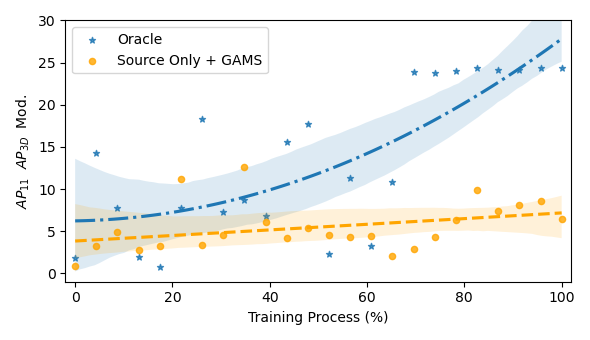}&
        \includegraphics[width=0.45\linewidth]{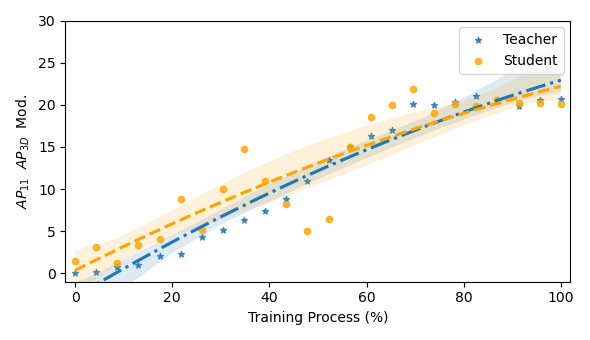}\\
        \scriptsize{(a) Oracle \textit{v.s.} Source Only + GAMS} & ~~~\scriptsize{(b) STMono3D Teacher \textit{v.s.} Student} \\
    \end{tabular}}
    \caption{Performance comparision. (a) Oracle \textit{v.s.} Source Only with GAMS: While the Oracle performance progressively improves, the Source Only model suffers from a drasical performance fluctuation. (b) Mean Teacher \textit{v.s.} Student on the target domain: Not only does the teacher model outperforms the student at the end of the training phase, its performance curve is also smoother and more stable.} 
    \label{fig::teacher}
\end{figure}

\subsubsection{Comparison of Self-Training Paradigm.}
We compare our STMono3D with other commonly used self-training paradigms (\textit{i.e.}, Naive ST) in Tab.~\ref{tab::ablation_teacher}. While the GAMS helps the Naive ST teacher generate effective pseudo labels on the target domain to boost UDA performance, our STMono3D still outperforms it by a significant margin. One of the primary concerns lies in low-quality pseudo labels caused by the domain gap. Moreover, as shown in Fig.~\ref{fig::teacher}(a), while the performance of Oracle improves progressively, the Source Only model on the target domain suffers from a performance fluctuation. It is also troublesome to choose a specific and suitable model from immediate results to generate pseudo labels for the student model. 

In terms of our STMono3D, the whole framework is trained in an end-to-end manner. The teacher is a temporal ensemble of student models at different time stamps. Fig.~\ref{fig::teacher}(b) shows that our teacher model is much more stable compared with the ones in Naive ST and has a better performance than the student model at the end of the training phase, where the teacher model starts to generate more predictions over the filtering score threshold. This validates our analysis in Sec.~\ref{subsec::stte} that the mean teacher paradigm provides a more effective teacher model for pseudo label generation. Tab.~\ref{tab::ablation_ema} demonstrates the effectiveness of the EMA of STMono3D. The performance significantly degrades when the EMA is disabled, and the model is easily crashed during the training stage. Moreover, since the model is simultaneously trained by data from both domains, our STMono3D can still preserve the knowledge from the source domain, which means a more powerful generalization capability. As shown in Tab.~\ref{tab::ablation_teacher}, STMono3D achieves even better results compared with Oracle models trained on the source domain.

\subsubsection{Effective of Quality-Aware Supervision.}
We study the effects of different applied loss terms of the proposed QAS strategy. Generally, the loss terms of Mono3D can be divided into two categories: ($i$) $\mathcal{L}_{cls}$ containing the object classification loss and attribute classification loss, and ($ii$) $\mathcal{L}_{reg}$ consisting of the location loss, dimension loss, and orientation loss. We separately apply the QAS on these two kinds of losses and report the corresponding results in Tab.~\ref{tab::ablation_QAS}. Interestingly, utilizing the confidence score from the teacher to reweight the $\mathcal{L}_{reg}$ cannot improve the model performance. We speculate it is caused by a loose correlation between the IoU score and localization quality (see yellow or blue line in Fig.~\ref{fig::score-iou}(a)), which is in line with the findings in LiDAR-based method~\cite{yang2021st3d}. However, we find QAS is more applicable for the $\mathcal{L}_{cls}$, where the model performance increases about 20.6\% $AP_{3D}$, which indicates the effectiveness of our proposed QAS strategy. It is intuitively reasonable since the score of pseudo labels itself is used to measure the confidence of predicted object classification. Such an instance-level reweighting strategy can help the model better handle low-quality pseudo labels as discussed in Sec.~\ref{subsec::qas}.

\subsubsection{Effective of Crucial Training Strategies.}
\begin{figure}[t]
    \centering
    \footnotesize
    \setlength{\tabcolsep}{14pt}
    \scalebox{0.95}{%
    \begin{tabular}{cc}
        \includegraphics[width=0.36\linewidth]{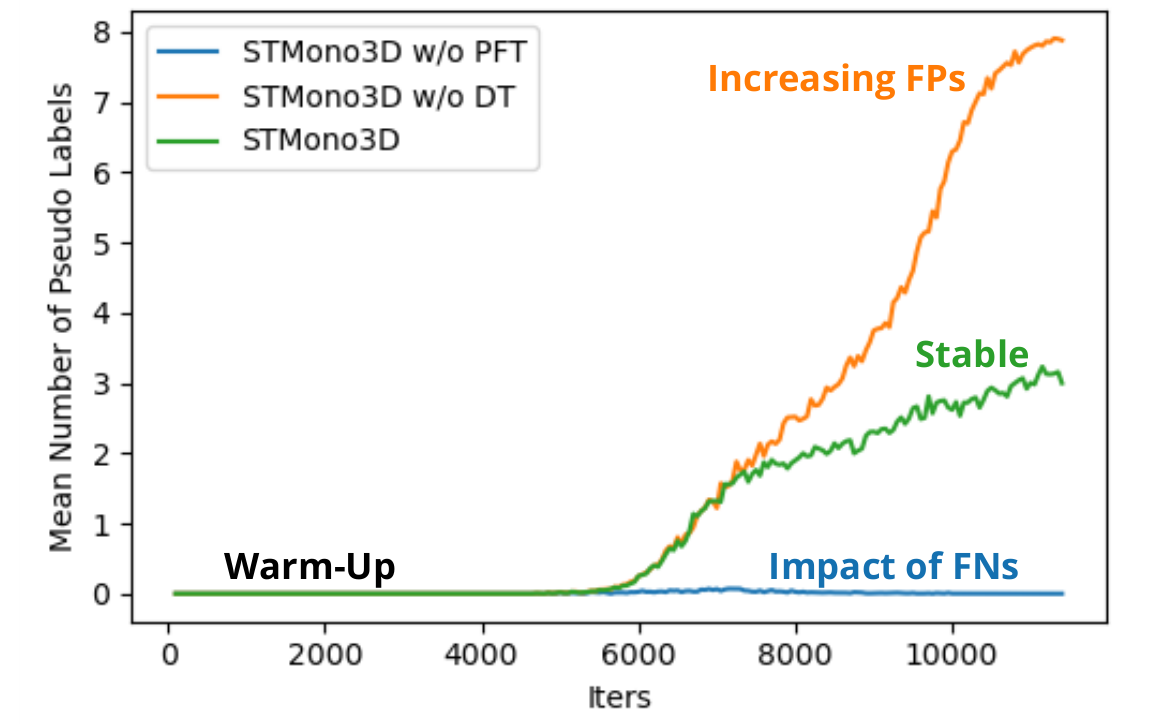}&
        \includegraphics[width=0.36\linewidth]{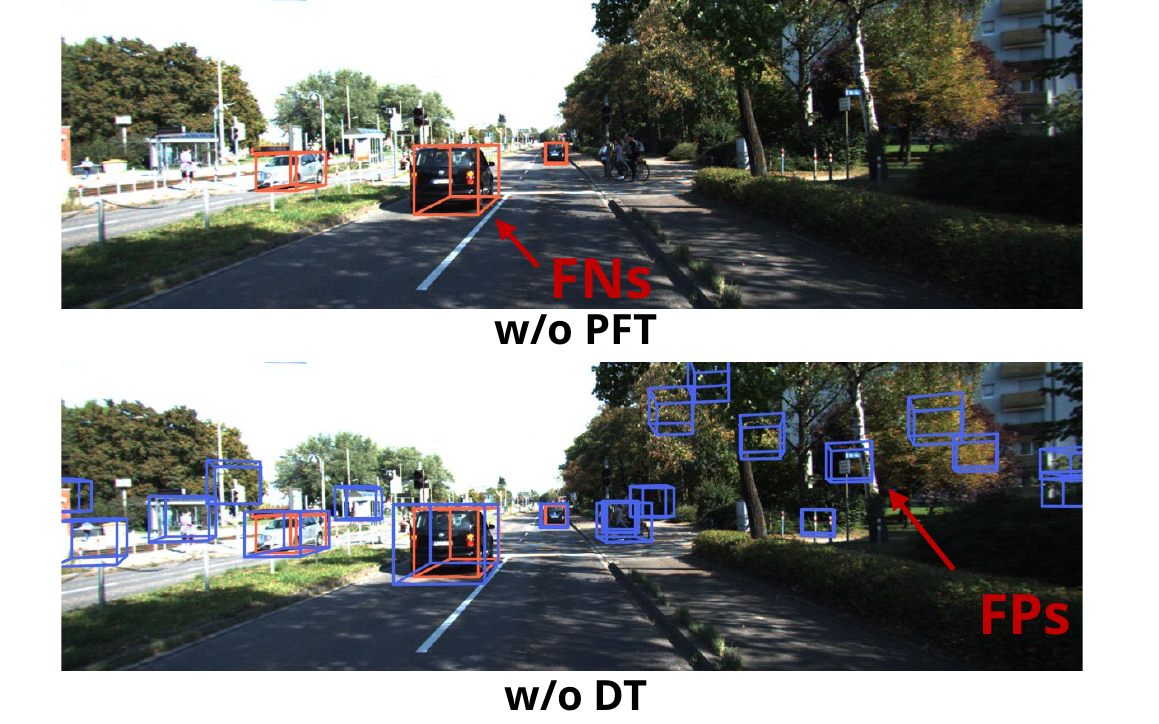}\\
        \scriptsize{(a) Num. of pseudo labels during training} & \scriptsize{(b) Visualization examples} \\
    \end{tabular}}
    \vspace{-0.15cm}
    \caption{Effects of the proposed DFT and DT. (a) Correlation between the average of the number of pseudo labels and training iters. (b) Examples of harmful FN and FP pseudo labels caused by disabling DFT and DT, respectively.} 
    \label{fig::abl_cts}
\end{figure}

We then further investigate the effectiveness of our proposed PFT and DT strategies. We first present the ablation results in Tab.~\ref{tab::ablation_CTS}. When we disable the strategies, model performance suffers from drastic degradations, where $AP_{3D}$ drops 64.3\%. The results demonstrate they are crucial strategies in STMono3D. As shown in Fig.~\ref{fig::abl_cts}(a), we also present the influence of them in a more intuitive manner. If we disable the PFT, the model will be severely impaired by the numerous FN predcitions (shown in Fig.~\ref{fig::abl_cts}(b) top) in the warm-up stage, leading to a failure to recognize objects in the following training iterations. On the other hand, for the teacher model w/o DT, the number of predictions abruptly increases at the end of training process, introducing more FPs predictions (shown in Fig.~\ref{fig::abl_cts}(b) down) that are harmful to the model perfomance. When jointly utilizing both the strategies, the number of pseudo labels stably increases, which means the detection capability of the model is gradually enhanced on the target domain.

\begin{table}[t]
    \centering
    \caption{Ablation study of PFT and DT.}
    \scalebox{0.8}{%
    \begin{tabular}{|cc|ccc|ccc|ccc|ccc|}
        \hline
        \multicolumn{2}{|c|}{\textbf{Nus$\rightarrow$K}} & \multicolumn{6}{c|}{$AP_{11}$} & \multicolumn{6}{c|}{$AP_{40}$} \\\hline
        \multirow{2}{*}{PFT} & \multirow{2}{*}{DT} &\multicolumn{3}{c|}{$AP_{BEV}$ $\text{IoU}\geqslant0.5$} & \multicolumn{3}{c|}{$AP_{3D}$ $\text{IoU}\geqslant0.5$} & \multicolumn{3}{c|}{$AP_{BEV}$ $\text{IoU}\geqslant0.5$} & \multicolumn{3}{c|}{$AP_{3D}$ $\text{IoU}\geqslant0.5$} \\
        & & Easy & Mod. & Hard & Easy & Mod. & Hard & Easy & Mod. & Hard & Easy & Mod. & Hard \\ 
        \hline
        & & 13.57 & 11.33 & 10.31 & 9.10 & 7.80 & 7.00 & 12.36 & 9.42 & 8.03 & 7.82 & 5.82 & 5.08 \\
        $\surd$& & 19.59& 16.00& 14.35 & 15.96& 13.15& 12.23 & 13.44& 9.76& 7.90 & 9.23 & 6.52 & 5.13 \\
        & $\surd$& 18.90& 16.57& 15.75& 15.15& 13.73& 12.85&12.74& 10.35& 9.42& 8.41& 6.81& 5.96 \\
        \cellcolor{LightCyan}$\surd$& \cellcolor{LightCyan}$\surd$& \cellcolor{LightCyan}35.63 & \cellcolor{LightCyan}27.37 & \cellcolor{LightCyan}23.95 & \cellcolor{LightCyan}28.65 & \cellcolor{LightCyan}21.89 & \cellcolor{LightCyan}19.55 & \cellcolor{LightCyan}31.85 & \cellcolor{LightCyan}22.82 & \cellcolor{LightCyan}19.30 & \cellcolor{LightCyan}24.00 & \cellcolor{LightCyan}16.85 & \cellcolor{LightCyan}13.66 \\
        \hline
    \end{tabular}
    }
\label{tab::ablation_CTS}
\vspace{-0.2cm}
\end{table}

\section{Conclusion}

In this paper, we have presented STMono3D, a meticulously designed unsupervised domain adaptation framework tailored for monocular 3D object detection task. We investigate that the depth-shift caused by the geometry discrepancy of domains leads to a drastic performance degradation when cross-domain inference. To alleviate the issue, we leverages a teacher-student paradigm for pseudo label generation and propose quality-aware supervision, positive focusing training and dynamic threshold to handle the difficulty in Mono3D UDA. Extensive experimental results demonstrate the effectiveness of STMono3D. In future work, we would like to explore temporal consistency to boost UDA performance.


\appendix
\section{Appendix}

\section{Dataset Comparisons}
\begin{figure}[t]
    \centering
    \footnotesize
    \scalebox{0.98}{%
    \begin{tabular}{c}
        \includegraphics[width=1\linewidth]{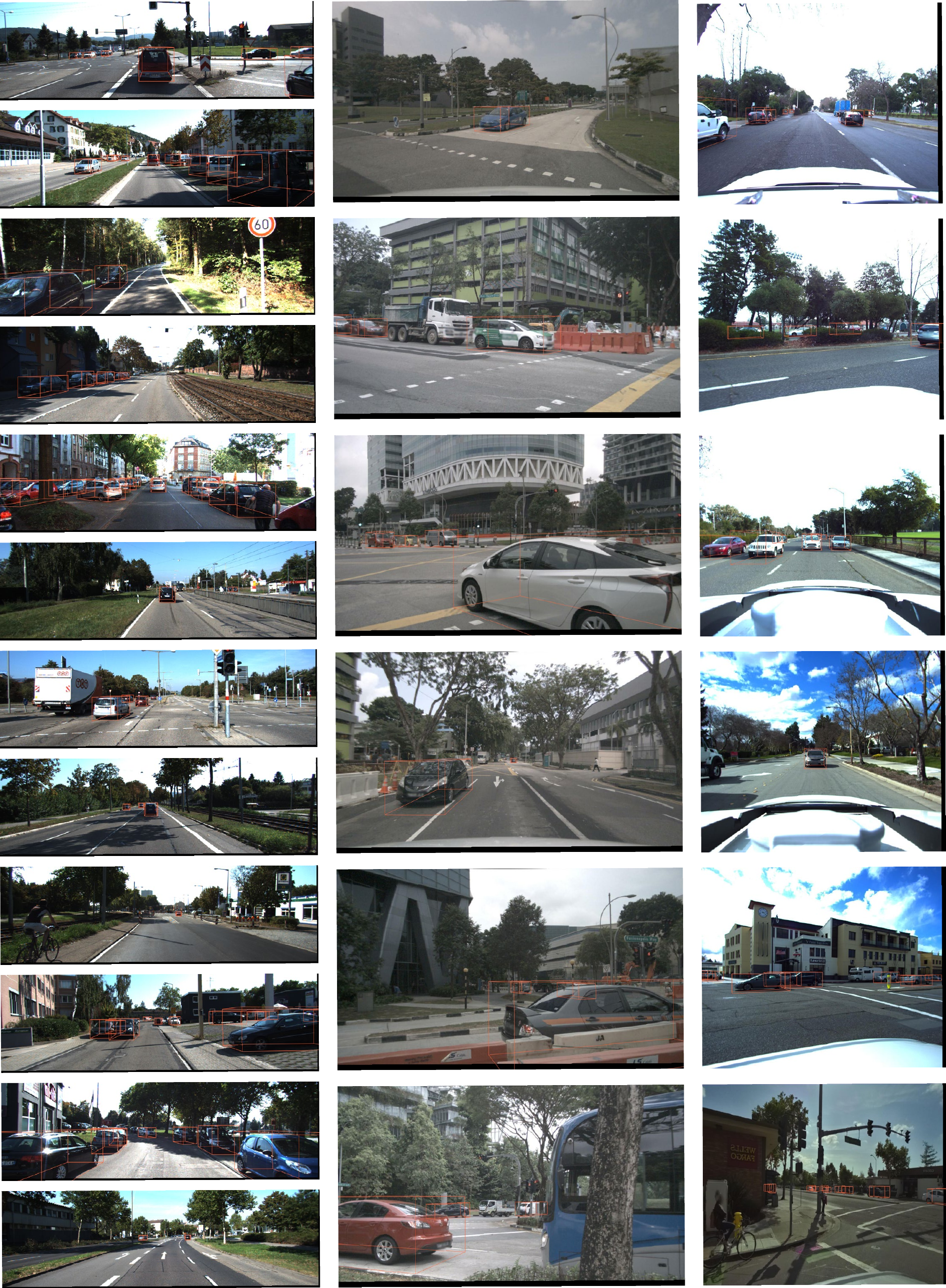}\\
        \hspace{0.04\linewidth}KITTI\hspace{0.27\linewidth} NuScense\hspace{0.23\linewidth} Lyft \\
    \end{tabular}}
    \vspace{-0.15cm}
    \caption{Dataset visualizations with ground-truth labels.}
    \label{fig::dataset-compare}
\end{figure}

To provide more intuitive comparisons among different datasets (\textit{e.g.}, KITTI~\cite{geiger2012kitti}, NuScense~\cite{caesar2020nusc} and Lyft~\cite{kesten2019lyft}), we present images with projected ground-truth labels in Fig.~\ref{fig::dataset-compare}. One can easily observe cameras utilized in these datasets have different parameters, which are reflected in the image resolutions, FOV, \textit{etc}. This work focuses on designing a general Mono3D UDA framework and solving the severe depth-shift caused by misaligned camera intrinsic parameters, which is the most crucial problem in Mono3D UDA. However, there are still numerous unsolved issues such as different image color styles, various distributions of object dimensions, different distributions of object depth, \textit{etc}. Our proposed STMono3D can be a well-developed baseline for future research.

\section{Visualizations of Pseudo Labels}
Here, we present more visualizations of pseudo labels generated by the teacher model during the training stage. The images show the depth-shift issue caused by the misalignment of camera parameters can be well-solved. The reasonable pseudo labels provide regular supervision on the target domain and achieve Mono3D UDA in a teacher-student paradigm. In addition, we can find there is still a slight error of prediction locations or dimensions that can be improved by further development of Mono3D methods and enhancement of the UDA algorithms. There is still tremedous room for improvement of the Mono3D UDA.

\section{Detailed Training Settings}
In this section, we introduce more detailed training settings. As for the model, we follow the basic config provided in MMDetection3D~\cite{mmdet3d2020}.The only modification lies in the scaling of predicted object depth based on the pixel size (GAMS introduced in our paper). We then summary all the runtime settings in Tab.~\ref{tab::settings}, including the number of interations, training schedule, threshold changing, \textit{etc}.

\begin{figure}[t]
    \centering
    \footnotesize
    \scalebox{0.98}{%
    \begin{tabular}{c}
        \includegraphics[width=1\linewidth]{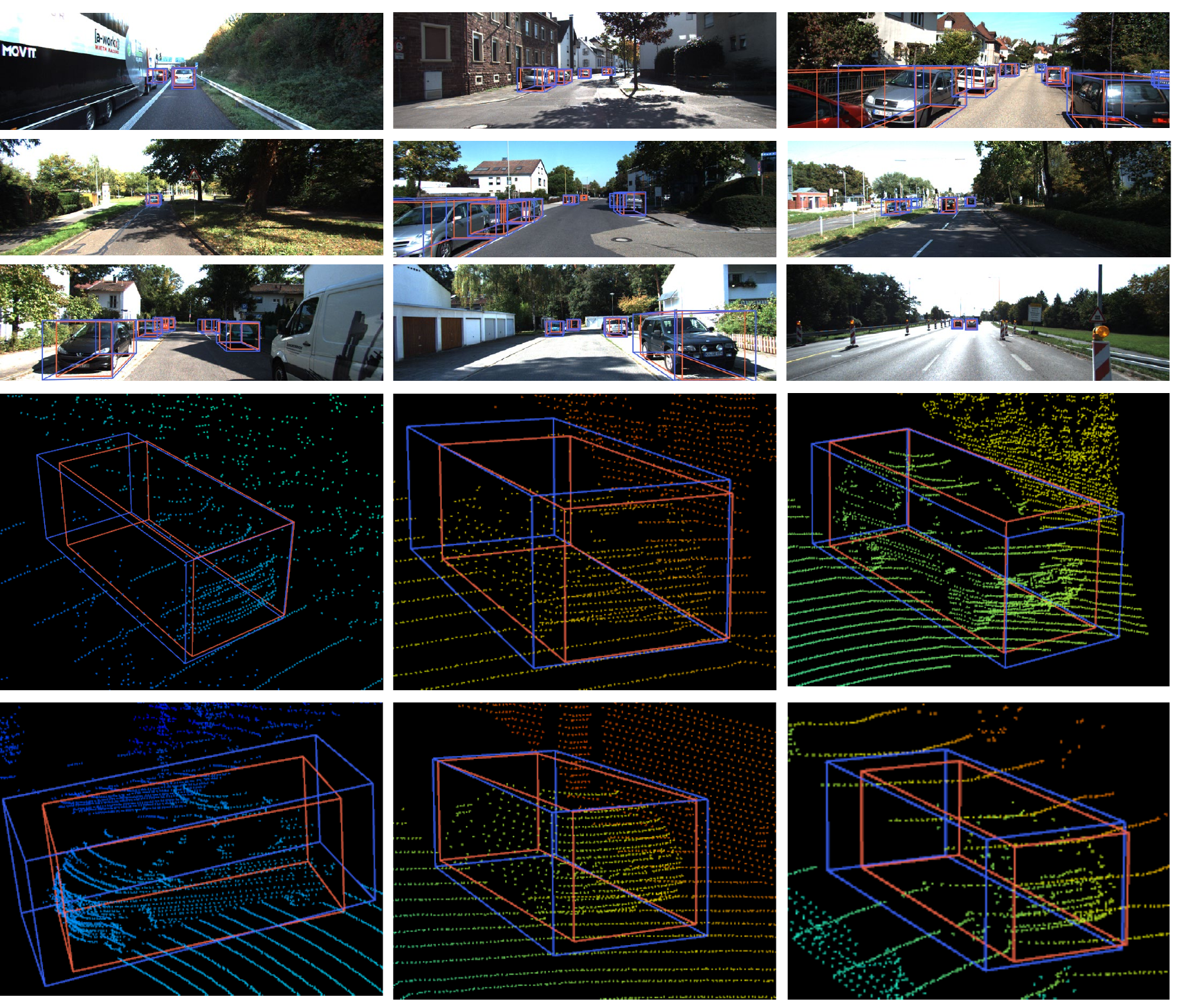}\\
    \end{tabular}}
    \vspace{-0.15cm}
    \caption{Visualizations of pseudo labels generated by the teacher model.}
    \label{fig::pseudo-labels}
\end{figure}

\clearpage
\begin{table}
    \centering
    \caption{Detailed training settings.}
    \label{tab::settings}
    \begin{minipage}{0.48\columnwidth}
        \centering
        \scalebox{0.9}{%
        \begin{tabular}{c|c}
            \multicolumn{2}{c}{\textbf{Schedule}}\\
            \hline
            number of iters& 880 $\times$ 13\\
            learning rate policy & step \\
            warmup type & linear \\
            warmup iters & 500\\
            warmup ratio & 1/3 \\
            step & [880 $\times$ 8, 880 $\times$ 11]
        \end{tabular}}
    \end{minipage}~~~~
    \begin{minipage}{0.48\columnwidth}
        \centering
        \scalebox{0.9}{%
        \begin{tabular}{c|c}
            \multicolumn{2}{c}{\textbf{Optimizer}}\\
            \hline
            optimizer type & ~~~SGD~~~ \\
            learning rate  & 0.002 \\
            gradient clip & True \\
            batchsize per GPU & 4\\
            number of GPUs & 8 \\
            source:domain samples per bs & 1:1\\
        \end{tabular}}
    \end{minipage}
    \\[12pt]
    \begin{minipage}{0.48\columnwidth}
        \centering
        \scalebox{0.9}{%
        \begin{tabular}{c|c}
            \multicolumn{2}{c}{\textbf{Mean Teacher Paradigm}}\\
            \hline
            momentum & ~~~0.999~~~ \\
            interval & 1 \\
            warmup & 0 \\
            ~increasing thr per step~ & 0.005 \\
            start iter & 880$\times$8 \\
            stop iter & 880$\times$10 \\
        \end{tabular}}
    \end{minipage}
    \begin{minipage}{0.48\columnwidth}
        \centering
        \scalebox{0.9}{%
        \begin{tabular}{c|c|c}
            \multicolumn{3}{c}{\textbf{Inference Settings (KITTI)}}\\
            \hline
            \multirow{4}{*}{~~student~~} & nms pre & ~~~100~~~ \\
            &nms thr & 0.05 \\
            &score thr & 0.001 \\
            &max per img & 20 \\
            \hline
            teacher & score thr (only diff.) & 0.35  \\
        \end{tabular}}
    \end{minipage}
    \\[12pt]
    \begin{minipage}{0.98\columnwidth}
        \centering
        \scalebox{0.9}{%
        \begin{tabular}{c|c|c}
            \multicolumn{3}{c}{\textbf{Strong Data Augmentation (type/prob/details)}}\\
            \hline
            ~~~RandomFlip3D~~~ & ~~~~~0.5~~~~~ & ~~~~~~~~horizontal~~~~~~~~\\
            \hline
            Mono3DResize & 1 & 
            \begin{tabular}{@{}c@{}}
                (1600, 840) (1600, 900)\\ 
                (1600, 960) (1600, 1020)\\
                (1600, 1080) (1600, 1140)\\ 
                (1600, 1200) (1600, 1260)\\
                (1540, 840) (1480, 780)\\
                (1420, 720) (1380, 680)\\
                (1660, 960) (1720, 1020)\\
                (1800, 1080) (1880, 1140)
            \end{tabular}\\
            \hline
            \multirow{8}{*}{OneOf} & \multirow{8}{*}{1} &Identity\\
            &&AutoContrast\\
            &&RandEqualize\\
            &&RandSolarize\\
            &&RandColor\\
            &&RandContrast\\
            &&RandBrightness\\
            &&RandSharpness\\
            &&RandPosterize\\
            \hline
            RandErase & 1 & 
            \begin{tabular}{@{}c@{}}
                size=[0, 0.2]\\ 
                n blocks=(1, 5)\\
                squared=True\\
            \end{tabular}\\
        \end{tabular}}
    \end{minipage}
    \\[12pt]
    \begin{minipage}{0.98\columnwidth}
        \centering
        \scalebox{0.9}{%
        \begin{tabular}{c|c|c}
            \multicolumn{3}{c}{\textbf{Weak Data Augmentation (type/prob/details)}}\\
            \hline
            ~~~RandomFlip3D~~~ & ~~~~~0.5~~~~~ & ~~~~~~~~horizontal~~~~~~~~\\
            \hline
            Mono3DResize & 1 & 
            \begin{tabular}{@{}c@{}}
                (1600, 840) (1600, 900)\\ 
                (1600, 960) (1600, 1020)\\
                (1600, 1080) (1600, 1140)\\ 
                (1600, 1200) (1600, 1260)\\
                (1540, 840) (1480, 780)\\
                (1420, 720) (1380, 680)\\
                (1660, 960) (1720, 1020)\\
                (1800, 1080) (1880, 1140)
            \end{tabular}\\
        \end{tabular}}
    \end{minipage}
\end{table}

\clearpage
%
%
\bibliographystyle{splncs04}
\bibliography{egbib}

\end{document}